**Important notice**

1. This is the preprint of the dataset descriptor paper:

   Gupta, R., Vishwanath, A., and Yang, Y. (2022). COVID-19 Twitter Dataset with Latent Topics, Sentiments and Emotions Attributes, *arXiv preprint*: https://doi.org/10.48550/arXiv.2007.06954

2. The data files described in this paper are available for download at OpenICPSR's COVID-19 Data Repository:

   Gupta, R., Vishwanath, A., and Yang, Y. (2022), COVID-19 Twitter Dataset with Latent Topics, Sentiments and Emotions Attributes, Ann Arbor, *MI: Inter-university Consortium for Political and Social Research* [distributor], 2022-06-25. https://doi.org/10.3886/E120321V12

3. The latest dataset version (V12, June 2022) has the following main updates: a) Full data coverage extended to cover 28 January 2020 – 1 June 2022 (2 years and 4 months), b) Country-specific CSV files download covers 30 representative countries, c) Added new vaccine-related data covering from 3 November 2021 to 1 June 2022 (8 months), d) an updated discussion on the dataset's usage.

4. The dataset license can be found in the OpenICPSR download folder. Essentially, the dataset license is based on the CC BY-NC 2.0 template and considers the need to be consistent with Twitter's terms of service, as the dataset is built upon the content provided by Twitter's standard API.

   Hence, you should read and agree with Twitter's Terms of Service, Privacy Policy, Developer Agreement, and Developer Policy if you intend to use the dataset. The user should also read the restricted uses from Twitter to avoid using the dataset for any potentially inappropriate use.

5. For inquiries, please contact the corresponding author via email.

# COVID-19 Twitter Dataset with Latent Topics, Sentiments and Emotions Attributes

Raj Kumar Gupta[1], Ajay Vishwanath[1,2], Yinping Yang[1]

[1] Affective Computing Group, Social and Cognitive Computing Department, Institute of High Performance Computing, Agency for Science, Technology and Research (A*STAR), Singapore

[2] Centre for Artificial Intelligence Research, University of Agder, Norway

Corresponding author: Yinping Yang (yangyp@ihpc.a-star.edu.sg)

## Abstract
This paper describes a large global dataset on people's discourse and responses to the COVID-19 pandemic over the Twitter platform. From 28 January 2020 to 1 June 2022, we collected and processed over 252 million Twitter posts from more than 29 million unique users using four keywords: "corona", "wuhan", "nCov" and "covid". Leveraging probabilistic topic modelling and pre-trained machine learning-based emotion recognition algorithms, we labelled each tweet with seventeen attributes, including a) ten binary attributes indicating the tweet's relevance (1) or irrelevance (0) to the top ten detected topics, b) five quantitative emotion attributes indicating the degree of intensity of the valence or sentiment (from 0: extremely negative to 1: extremely positive) and the degree of intensity of fear, anger, sadness and happiness emotions (from 0: not at all to 1: extremely intense), and c) two categorical attributes indicating the sentiment (very negative, negative, neutral or mixed, positive, very positive) and the dominant emotion (fear, anger, sadness, happiness, no specific emotion) the tweet is mainly expressing. We discuss the technical validity and report the descriptive statistics of these attributes, their temporal distribution, and geographic representation. The paper concludes with a discussion of the dataset's usage in communication, psychology, public health, economics, and epidemiology.

## Background & Summary
The 2019 Coronavirus Disease (COVID-19) was first officially reported as an acute respiratory infection caused by an unknown virus in Wuhan city, Hubei province in China, on 31 December 2019. According to the World Health Organization (WHO) 's COVID-19 situation dashboard[1], as of 17 June 2022, the disease has infected 535,863,950 people worldwide and claimed 6,314,972 lives [1].

The pandemic presents complex and evolving problems that warrant multidisciplinary research and a globally concerted effort. The complexity comes from the disease itself and the surge of the medical, scientific, social, behavioural, and economic issues that the disease has brought about. These issues include reports on daily counts of new cases and mortality rates, scientific discoveries, government responses, news reporting of social behaviours such as panic buying and food hoarding, impact on businesses and economic outlook, and changes in people's everyday lives. The problems are multi-faceted and unprecedented. There is a growing recognition of the need for multidisciplinary research efforts to support the COVID-19 pandemic response, including disciplines such as social and behavioural science [2] and mental health science [3].

---
[1] https://covid19.who.int

Twitter is a popular microblogging site widely used by Internet users. According to Statista, as of the first quarter of 2022, Twitter had 229 million active users worldwide, an increase of over 50 per cent from the 152 million users in the fourth quarter of 2019[2] [4]. Twitter provides the research community with a rich source of information about when, where, and what people have to say in their posts (known as "tweets") through its free, publicly accessible standard application programming interface (API) service. However, the raw tweet content is mainly in textual format and is not readily analysable. When there are many tweets, it takes a significant time to accurately extract information about people's concerns, feelings, and emotions for analysts and researchers to process and analyse for in-depth patterns and insights.

This paper describes the data collection and processing methods that enable the development of the full dataset we released for academic research use since its first version in July 2020. With tweet-by-tweet labelled data systematically extracted and made available to the research communities, our purpose is to enable more researchers to perform in-depth investigations in all possible research-worthy areas, such as discovering the correlational patterns between other variables such as government measures and communication of these measures, demographics, economic indicators, and epidemiological markers. Figure 1 presents the schematic structure of the data.

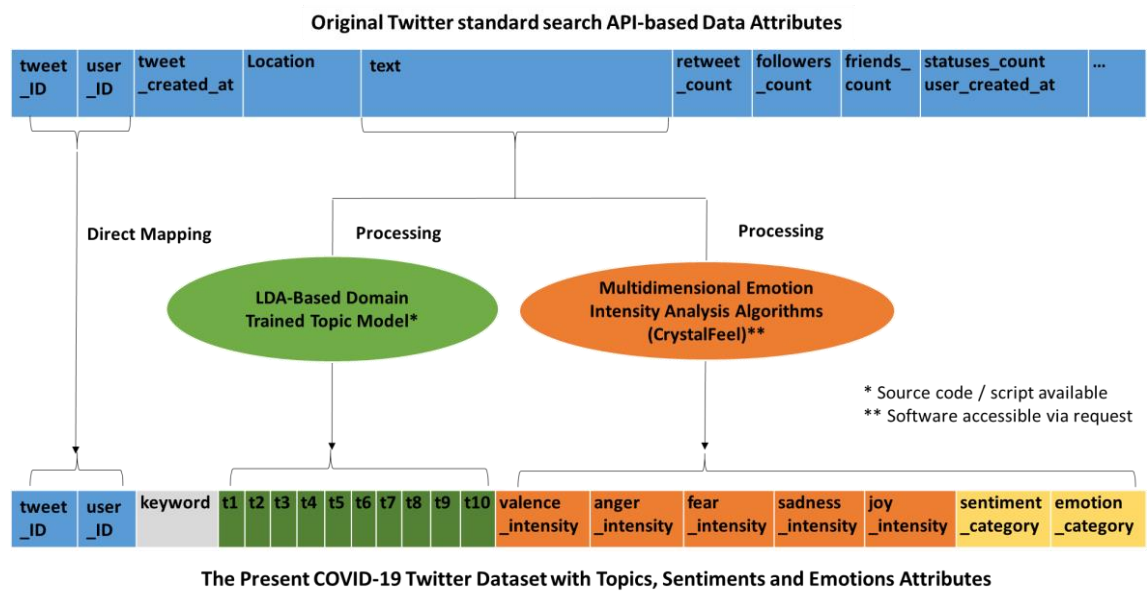

**Figure 1. Overview of the before-after data structure and the processing methods applied**

As early as the first half of 2020, a few studies have leveraged Twitter data for COVID-19 research (e.g., [5, 6, 7]). Twitter has also opened up real-time, full-fidelity data streams[4] related to COVID-19 tweets since late April 2020 [8]. Researchers have also systematically collected and shared COVID-19-related tweets datasets for open research use [9, 10, 11]. However, to the best of our knowledge, no existing datasets provide a research resource with rich, semantically, and psychologically meaningful attributes surrounding the topics, sentiments, and emotions content of the tweets. In contrast, the dataset described in this paper provides a novel, tweet-by-tweet level tagging of the topic clusters that a tweet is semantically related to, the sentiment the tweet is expressing, and the emotional properties, in terms of the intensity of the emotion as well as the type of the emotion, associated with each of the tweets.

---
[2] https://www.statista.com/statistics/970920/monetizable-daily-active-twitter-users-worldwide
[4] https://developer.twitter.com/en/docs/labs/covid19-stream/overview

The subsequent sections of the paper describe the data collection and processing methods with a focused interest in tracking and understanding the latent topics, sentiments, and emotions surrounding the COVID-19 pandemic. We applied natural language processing (NLP) techniques, specifically statistical topic clustering techniques, that detect tweets surrounding similar topic clusters. We also applied pre-trained machine learning algorithms (CrystalFeel) to tag each tweet with a sentiment valence (unpleasantness/pleasantness) and intensity scores of four different emotions – fear, anger, happiness, and sadness.

## Methods
**Collection of raw Twitter data related to COVID-19 and vaccine**
We set up our data collection app in early February 2020 by querying Twitter's standard search API[5] [13]. We first used three keywords "corona", "wuhan" (many people refer to the virus as "wuhan virus" at the initial stages before the official name was announced), and "nCov" (WHO first named the virus as "2019-nCov"). On 11 February 2020, upon WHO's officially renaming of the disease as "COVID-19", we added "covid" as a new search keyword.

As vaccination is one of the key measures to fight against the COVID-19 virus, we also started to collect vaccine-related tweets in late October 2021. We used nine keywords "vaccine", "pfizer", "moderna", "covaxin", "astrazeneca", "covishield", "janseen", "sinova", and "sinopharm".

We focused on English tweets and used the language filter with the Twitter API. Simple sharing of these tweets (i.e., retweets) is not collected for the dataset to ensure the conciseness of the data.

The Twitter API returns the tweet text content with a rich range of attributes. For example, we were able to download the following 12 attributes in our local database.

- **tweet_ID**: The unique identifier for this tweet
- **tweet_created_at**: The UTC time when this tweet was created
- **text**: The actual UTF-8 text of the tweet
- **retweet_count**: Number of times this tweet has been retweeted
- **favorite_count**: Number of times this tweet has been liked
- **hashtag_text**: Name of the hashtag, minus the leading '#' character
- **user_ID**: The unique identifier for the user who created the tweet
- **location**: The user-defined location in this account's profile
- **followers_count**: Number of followers this account currently has
- **friends_count**: Number of users this account is following
- **statuses_count**: Number of tweets (including retweets) issued by the user
- **user_created_at**: The UTC datetime that the user account was created on Twitter

**Processing for the topic attributes**
Although all the retrieved tweets are relevant to at least one of the four COVID-19-related keywords, many facets or subtopics have been covered in the tweets' "text" content. We applied an unsupervised topic clustering technique called Latent Dirichlet Allocation (LDA) to understand the subtopics. LDA is a probabilistic generative model which learns a multinomial distribution of latent topics in a given document [14]. The advantage of LDA is that it is independent of the corpus size, making it algorithmically efficient to learn topic clusters within a corpus with many tweets such as ours.

---
[5] https://developer.twitter.com/en/docs/tweets/search/overview/standard

First, we pre-processed each raw tweet by converting it to ASCII characters, removing accented characters, forming bigrams and trigrams, filtering out stop words (including most rare and most frequent words), and performing text tokenization. These pre-processed tweets were then converted into a bag-of-words (BoW) corpus. The training data's date range is 28 January 2020 to 27 May 2020, consisting of 51 million tweets.

Next, we randomly sampled 1% of the BoW corpus and trained an LDA model whose inference was performed using online variational Bayes [15]. Using the trained LDA-based topic model, we obtained 100 topic clusters. The following list illustrates the top ten topics detected in the dataset (e.g., "topic 1") and the ten most representative words associated with each detected topic (e.g., "people, cases, new, deaths, time, china, realdonaldtrump, lockdown, trump" for "topic 1"), respectively.

- **topic 1**: people, cases, new, deaths, time, china, realdonaldtrump, lockdown, trump
- **topic 2**: health, help, people, need, think, vaccine, care, fight, support
- **topic 3**: pandemic, f**k, months, killed, question, wait, looks, trump, impact
- **topic 4**: pay, donate, lie, focus, song, gates, page, google, caused
- **topic 5**: florida, drink, named, nature, marketing, pr, ncdcgov, farmers, cr
- **topic 6**: rules, bed, drtedros, speaks, privacy, parliament, physicians, strength, joke
- **topic 7**: dies, pmoindia, ndtv, ai, narendramodi, mohfwindia, shoot, drharshvardhan, battle
- **topic 8**: ye, ke, behaviour, brought, hidden, yup, smell, zerohedge, odds
- **topic 9**: excuse, humanity, salary, wind, gtgt, rats, ice, beard, mosque
- **topic 10**: internet, allah, teacher, dance, el, rona, weed, crush, fk

Lastly, for each tweet in the entire dataset, we assigned a relevance label ("1" or "0") using the trained LDA model based on the contribution of each topic (over a total of 100 topic clusters) to the tweet ("1" indicates if the contribution is > 1%, where "0" indicates otherwise).

Table 1 shows example tweets that are tagged with the corresponding topic clusters, respectively. The first ten examples show tweets that are solely relevant to each of the ten topic clusters. The last example shows that a tweet can be relevant to multiple topic clusters simultaneously.

**Table 1. Examples of the tweets and their corresponding topic attributes**

| Example tweet text | t1 | t2 | t3 | t4 | t5 | t6 | t7 | t8 | t9 | t10 |
|---|---|---|---|---|---|---|---|---|---|---|
| "Remember when doja cat said corona was just the flu" | 1 | 0 | 0 | 0 | 0 | 0 | 0 | 0 | 0 | 0 |
| "STAY SAFE from CORONAVIRUS There is currently no antiviral treatment or vaccine to prevent Coronavirus (COVID-19)" | 0 | 1 | 0 | 0 | 0 | 0 | 0 | 0 | 0 | 0 |
| "No. It is a part of One Belt One Road. #ChinaVirus WHO says Wuhan coronavirus outbreak is not yet a pandemic" | 0 | 0 | 1 | 0 | 0 | 0 | 0 | 0 | 0 | 0 |
| "Bill Gates: This is how long it may take before Americans can be completely safe from COVID-19" | 0 | 0 | 0 | 1 | 0 | 0 | 0 | 0 | 0 | 0 |
| "Florida's COVID-19 website guru blasts bosses, hints at data suppression" | 0 | 0 | 0 | 0 | 1 | 0 | 0 | 0 | 0 | 0 |
| "Municipality closes 336 shops in all six governorates for violating corona rules" | 0 | 0 | 0 | 0 | 0 | 1 | 0 | 0 | 0 | 0 |
| "@Naveen_Odisha @narendramodi @MoHFW_INDIA @PrakashJavdekar @drharshvardhan @dr_arunsahoo I request to postpone our…" | 0 | 0 | 0 | 0 | 0 | 0 | 1 | 0 | 0 | 0 |
| "Previously its DAM funds. now ita corona fund. pochna sirf ye tha k DAM fund kaha tk pohancha" | 0 | 0 | 0 | 0 | 0 | 0 | 0 | 1 | 0 | 0 |
| "apparently child abuse cases are going up and the excuse is that parents are stressed over covid-19. im sorry but" | 0 | 0 | 0 | 0 | 0 | 0 | 0 | 0 | 1 | 0 |
| "today i found out weed kills the corona virus and i ain't been worried since" | 0 | 0 | 0 | 0 | 0 | 0 | 0 | 0 | 0 | 1 |
| "PICS: Built in 10 days – 1000 bed hospital opens to battle coronavirus in Wuhan All 4 Women" | 1 | 1 | 1 | 0 | 1 | 0 | 0 | 0 | 0 | 0 |

**Processing for the sentiment and emotion intensity attributes**

As sentiments and emotions are subjective information embedded in the unstructured "text" content, extracting such information with targeted tools is necessary.

We used CrystalFeel[6] [16], a collection of five machine-learning-based algorithms to extract the sentiment and emotions scores on both quantitative and qualitative scales. The development of CrystalFeel involved training and experimental evaluations of features derived from affective lexicons, parts-of-speech, and word embeddings [17], using tweets manually annotated with ground truth values [18].

Table 2 shows five example tweets tagged with these five attributes: valence intensity, fear intensity, anger intensity, happiness intensity, and sadness intensity. The first example shows a tweet with a moderate (i.e., neither very negative nor very positive) "valence_intensity".

---

[6] CrystalFeel is accessible via https://socialanalyticsplus.net/crystalfeel

The other four examples show tweets with relatively high-intensity scores for happiness, anger, fear, and sadness dimensions.

Note that, in some instances, such as the fourth example ("Being higher risk of covid has me all over the place. Appt is in about 90 mins. Im scared, worried and anxious"), the intensity score may be exceeding 0-1 range, which indicates that these cases represented extreme intensities beyond the algorithms' original training samples.

**Table 2. Examples tweets and their corresponding sentiment valence and emotions intensity scores attributes**

| Example tweet text | valence_intensity | fear_intensity | anger_intensity | happiness_intensity | sadness_intensity |
|---|---|---|---|---|---|
| "Community hospital Bright Vision transfers all patients to make room for stable COVID-19 cases" | **0.505** | 0.444 | 0.391 | 0.334 | 0.423 |
| "Seeing this face for just a few minutes on her 97th birthday today made my heart so happy! #Greatful #COVID #Nana" | 0.930 | 0.177 | 0.143 | **0.822** | 0.209 |
| "To any fellow nationalist celebrating boris getting the corona virus are nothing less than an absolute cunt." | 0.318 | 0.520 | **0.665** | 0.260 | 0.495 |
| "Being higher risk of covid has me all over the place. Appt is in about 90 mins . Im scared, worried and anxious" | 0.231 | **1.074** | 0.542 | 0.174 | 0.670 |
| "When u cant handle shit any longer and u feel hopeless #COVID_19 starting to make my depression worse" | 0.076 | 0.836 | 0.617 | 0.077 | **0.940** |

**Processing for the sentiment category attribute**

To facilitate a more straightforward interpretation, we used the CrystalFeel algorithm's qualitative output "sentiment" for the dataset. The "sentiment" attribute is derived from the following logic from "valence_intensity" score.

```
//    # Initialize the sentiment category in a "neutral or mixed" class
1         sentiment = "neutral or mixed";
//    # Assign the sentiment category based on the degree of the valence intensity
2     if(valence_intensity <= 0.30):
3         sentiment = "very negative";
4     elif(valence_intensity < 0.48):
5         sentiment = "negative";
6     elif(valence_intensity > 0.70):
7         sentiment = "very positive";
8     elif(valence_intensity > 0.52):
9         sentiment = "positive";
```

Table 3 shows the five tweets examples tagged with their corresponding sentiment categories, qualitatively indicating the sentiment each tweet is mainly expressing.

**Table 3. Examples of the tweets data and their corresponding sentiment categories**

| Example tweet text | sentiment |
|---:|---:|
| "Seeing this face for just a few minutes on her 97th birthday today made my heart so happy! #Greatful #COVID #Nana" | very positive |
| "Community hospital Bright Vision transfers all patients to make room for stable COVID-19 cases" | neutral or mixed |
| "To any fellow nationalist celebrating boris getting the corona virus are nothing less than an absolute cunt. " | negative |
| "Being higher risk of covid has me all over the place. Appt is in about 90 mins . Im scared , worried and anxious" | very negative |
| "When u cant handle shit any longer and u feel hopeless #COVID_19 starting to make my depression worse" | very negative |

**Processing for the emotion category attribute**

The underlying emotion behind the sentiments carries more information than the overall valence or sentiment. We used CrystalFeel's "emotion" output to facilitate the interpretation of the COVID-19 tweets.

The "emotion" output was obtained using the following logic that leverages all the valence and emotions intensities scores from CrystalFeel's outputs. It first uses "valence_intensity" as the first-line criterion as this dimension has very high accuracy, i.e., 0.816 in terms of Pearson correlation with human-annotated ground truth values [16, 17]. It then uses the relative intensity comparing the three primary negative emotions, anger, fear and sadness to assign a corresponding dominant emotion category. Whenever there is a tie, i.e., a tweet may have identical intensity scores across fear, anger, and sadness, the emotion category is assigned as "anger", followed by "fear", and "sadness". The empirical basis is that anger intensity has the highest accuracy, i.e., 0.818 in terms of the Pearson correlation value [16, 17]. The following script describes the conversion logic:

```
//     # Initialize the sentiment category in a "no specific emotion" class
1              emotion = "no specific emotion";
//     # Assign the emotion category when valence intensity score exceeds 0.52
2      if(valence_intensity > 0.52):
3              emotion = "happiness";
//     # Assign the emotion category when valence intensity score falls below 0.48
4      elif(valence_intensity < 0.48):
5              emotion = "anger";
6      if((fear_intensity > anger_intensity) and (fear_intensity > = sadness_intensity )):
7              emotion = "fear";
8         elif((sadness_intensity > anger_intensity) and (sadness_intensity > fear_intensity)):
9              emotion = "sadness";
```

Table 4 shows the five tweet examples tagged with the dominant emotion categories.

**Table 4. Examples of the tweets data and their corresponding emotion category results**

| Example tweet text | emotion |
|---|---|
| "Being higher risk of covid has me all over the place. Appt is in about 90 mins . Im scared , worried and anxious" | fear |
| "To any fellow nationalist celebrating boris getting the corona virus are nothing less than an absolute cunt. " | anger |
| "Seeing this face for just a few minutes on her 97th birthday today made my heart so happy! #Greatful #COVID #Nana" | happiness |
| "When u cant handle shit any longer and u feel hopeless #COVID_19 starting to make my depression worse" | sadness |
| "Community hospital Bright Vision transfers all patients to make room for stable COVID-19 cases" | no specific emotion |

It is helpful to note that the conversion logic mentioned above is based on application assumptions where CrystalFeel is used for processing *short informal text* (e.g., tweets, Facebook posts, and comments), which is the case for the present Twitter COVID-19 dataset. The conversion thresholds are derived from heuristics and social media corpora we continuously monitor in our research (see more in [16]).

**Processing for timestamp attribute**
We processed the original "tweet_created_at" (in unix format) and obtained a "tweet_timestamp" attribute (in YYYY-MM-DD-HH-SS format). The timezone is maintained as UTC time.

**Processing for country/region attribute**
As our COVID-19 data collection is keyword-based, the tweets returned appear to come from users from different geographic regions worldwide. To facilitate the assessment of the geographic representativeness of the dataset, we converted the "location" attribute from the original Twitter results into a "country/region" attribute. This is done using GeoNames' cities15000 geographic database[7] [19], which contains a mapping between all cities with a population > 15,000 or capitals and a country code.

For example, the original location "Ontario, Canada" is converted to country/region code as "Canada", "India" is converted to "India", "Shanghai" is converted to "China", and "London" is converted to "United Kingdom".

If the location is indicated as "online", "The Entire Universe!" or left blank (i.e., no match can be found using the GeoNames database), the country_region is coded as "-", indicating that there is no country or region identifiable information associated with the tweet.

If the user does not indicate any location information, the country/region is maintained as an empty field.

## Data Records
The data records are constructed as comma-separated value (CSV) files.

---
[7] http://download.geonames.org/export/dump/cities15000.zip

As of 1 June 2022, our system collected a total of 252,600,524 tweets worldwide using the four COVID-19-related keywords. The system also collected a total of 13,505,737 tweets using the nine vaccine-related keywords.

- **Time coverage (COVID-19)**: 28 January 2020 – 1 June 2022 (2 years and 4 months)
- **Total number of tweets retrieved/number of rows (COVID-19):** 252,600,524
- **Total number of tweets' unique users (COVID-19):** 29,393,115
- **Time coverage (Vaccine)**: 3 November 2021 – 1 June 2022 (8 months)
- **Total number of tweets retrieved/number of rows (Vaccine):** 13,505,737
- **Total number of tweets' unique users (Vaccine):** 2,848,210
- **Geographic coverage**: Worldwide
- **Language:** English
- **Total number of columns/attributes/variables:** 22

All the data record files have at least the following three columns or attributes.

- **tweet_ID**: the unique identifier for this tweet
- **user_ID**: the unique identifier for the user
- **keyword**: one of the four keywords ("corona", "wuhan", "nCov" or "covid") we used to query the Twitter API, which returned the tweet

1. tweetid_userid_keyword_topics.csv

This file contains the entire tweets CSV file with the following ten attributes of the processed topic. The file is very large. We recommend that users use python+pandas to view and retrieve data records in this file.

- **t1**: A binary value of 0 or 1, where 0 – this tweet is not relevant to this topic; 1 – this tweet is relevant to this topic
- **t2**: A binary value of 0 or 1, where 0 – this tweet is not relevant to this topic; 1 – this tweet is relevant to this topic
- **t3**: A binary value of 0 or 1, where 0 – this tweet is not relevant to this topic; 1 – this tweet is relevant to this topic
- **t4**: A binary value of 0 or 1, where 0 – this tweet is not relevant to this topic; 1 – this tweet is relevant to this topic
- **t5**: A binary value of 0 or 1, where 0 – this tweet is not relevant to this topic; 1 – this tweet is relevant to this topic
- **t6**: A binary value of 0 or 1, where 0 – this tweet is not relevant to this topic; 1 – this tweet is relevant to this topic
- **t7**: A binary value of 0 or 1, where 0 – this tweet is not relevant to this topic; 1 – this tweet is relevant to this topic
- **t8**: A binary value of 0 or 1, where 0 – this tweet is not relevant to this topic; 1 – this tweet is relevant to this topic
- **t9**: A binary value of 0 or 1, where 0 – this tweet is not relevant to this topic; 1 – this tweet is relevant to this topic
- **t10**: A binary value of 0 or 1, where 0 – this tweet is not relevant to this topic; 1 – this tweet is relevant to this topic

It is useful to note that the current release includes topic information covering the time range of 28 January 2020 to 1 January 2021.

2. tweetid_userid_keyword_sentiments_emotions.csv

This file contains the entire tweets file with the processed seven sentiments and emotions attributes. The file is very large. We recommend that users use python+pandas to view and retrieve data records in this file.

- **valence_intensity**: A continuous variable ranging from 0 to 1, where 0 indicates that this text expresses extremely negative or unpleasant feelings, and 1 indicates that this text expresses extremely positive or pleasant feelings
- **fear_intensity**: A continuous variable ranging from 0 to 1, where 0 indicates that this text does not express the fear emotion at all, and 1 indicates that this text expresses an extremely high intensity of the fear emotion
- **anger_intensity**: A continuous variable ranging from 0 to 1, where 0 indicates that this text does not express the anger emotion at all, and 1 indicates that this text expresses an extremely high intensity of the anger emotion
- **happiness_intensity**: A continuous variable ranging from 0 to 1, where 0 indicates that this text does not express the happiness emotion at all, and 1 indicates that this text expresses an extremely high intensity of the happiness emotion
- **sadness_intensity**: A continuous variable ranging from 0 to 1, where 0 indicates that this text does not express the sadness emotion at all, and 1 indicates that this text expresses an extremely high intensity of the sadness emotion
- **sentiment**: A categorical variable that indicates the text mainly expresses one of the five sentiment classes: very negative (-2), negative (-1), neutral or mixed (0), positive (1), and very positive (2)
- **emotion**: A categorical variable that indicates the text mainly expresses one of the five emotion classes: fear (F), anger (A), sadness (S), happiness (H), and no specific emotion (NA)

In addition, the data file includes the following processed attributes.

- **Tweet_timestamp**: A timestamp in YYYY-MM-DD HH-MM-SS format (in UTC time) processed based on "time_created_at" retrieved from Twitter API
- **country/region**: A text field that indicates the country or region processed based on the "location" declared in the Twitter author's profile

Individual CSV files extracted for 30 representative countries are also included in the latest release.

**Hydrating other attributes**. In compliance with Twitter's content redistribution terms, our released dataset only contains two original Twitter data attributes: "tweet_ID" and "user_ID".

Users may use "tweet_ID" or "user_ID" to retrieve or "hydrate" the other attributes (such as the actual "text", "retweet_count", "location", "followers_count") through the standard search API from Twitter directly [13].

The following provides a simple step-by-step guide to hydrate Twitter data using Twitter standard search API and Python.

1) Request access to Twitter API via Twitter's developer site[8].
    a. Applying for a developer account on Twitter
    b. Choose between product tracks (we recommend Standard)

---

[8] https://developer.twitter.com/en/docs/twitter-api/getting-started/getting-access-to-the-twitter-api

c.  Get approval from Twitter
2) After getting the approval to use the API, a Project and App must be created which would have its designated API Key, API Key Secret, Access Token, Access Token Secret, and Bearer token.
3) If Python is the language of choice, we recommend using tweepy[9] to access Twitter data. A simple python implementation guide[10] provides further details.

Apart from the steps mentioned here, there are several third-party tools one could use to hydrate Twitter data, such as Hydrator[11] and twarc[12]. As they are third-party tools, users should carefully check the terms of use before deciding to use them.

## Technical Validation
**Overall raw tweets coverage**
As of 1 June 2022, our system has collected a total of 252,600,524 tweets worldwide using the four COVID-19-related keywords, with the first retrievable date being 28 January 2020.

Table 6 and Table 7 present the data overview. Most of the COVID-19-related tweets are retrieved based on the "covid" keyword, which returned 222,303,027 tweets, or 88.0%. On average, 12,296 COVID-related tweets were posted every hour, or 295,094 tweets every day. In total, 29,393,115 unique users posted these COVID-19-related tweets based on the "user_ID" attribute. Most of the vaccine-related tweets are retrieved based on the "vaccine" keyword, followed by "Pfizer" and "moderna". In total, 2,848,210 unique users posted these vaccine-related tweets based on the "user_ID" attribute.

**Table 6. COVID-19 Twitter data overview by keywords (28 January 2020 – 1 June 2022)**

| Keyword | Count of tweets collected | Share |
|---|---|---|
| corona | 25,302,298 | 10.0% |
| wuhan | 4,301,501 | 1.7% |
| ncov | 693,698 | 0.3% |
| covid | 222,303,027 | 88.0% |
| total | 252,600,524 | 100% |

**Table 7. Vaccine Twitter data overview by keywords (3 November 2021 – 1 June 2022)**

| Keywords | Count of tweets collected | Share |
|---|---|---|
| vaccine | 10,509,680 | 77.8% |
| pfizer | 1,880,577 | 13.9% |
| moderna | 710,963 | 5.3% |
| covaxin | 152,457 | 1.1% |
| astrazeneca | 122,433 | 0.9% |
| covishield | 42,255 | 0.3% |
| janssen | 36,703 | 0.3% |
| sinovac | 36,284 | 0.3% |
| sinopharm | 14,285 | 0.1% |
| total | 13,505,737 | 100% |

---

[9] https://www.tweepy.org
[10] https://github.com/ajvish91/covid_twitter_scripts/blob/master/hydrate_tweets.ipynb
[11] https://github.com/DocNow/hydrator
[12] https://github.com/DocNow/twarc

It is helpful to note the following limitations associated with the dataset.

1. Twitter's standard search API has a known limitation because it does not guarantee the retrieved tweets are exhaustive due to indexing and other reasons. In other words, the search API retrieves *relevant* but *not all* the tweets that match the search keywords.
2. We were able to retrieve only the first 144 characters of the tweet "text" from the Twitter standard search API, from 28 January 2020 to 18 March 2021. After 19 March 2021, we were able to retrieve and hence have processed full tweet content that may exceed 144 characters. The length of tweets may affect the topic and emotion analysis results before and after 18 March 2021. Re-processing is recommended for applications that require comparison before or after the date.
3. Data were missing for four days from 30 April 2022 to 3 May 2022, due to some service interruption from Twitter API.

**Validity of the processing methods**
**Topic identification**. The quality of the topic model was evaluated using metrics including perplexity and coherence scores based on suggestions from the literature [20]. We obtained the top ten topics, i.e., "t1", "t2", …, and "t10", that received relatively high coherence scores (c_v measure, mean = 0.575) from a model optimized by learning 100 topics and hyper-parameters α as a fixed normalized asymmetric Dirichlet prior (1/topic_number) and η = 0.909. We obtained ten topics out of 100 extracted from 500,000-odd data points, a 1% sample from the entire dataset.

Conceivably, training an LDA-based topic model with data from specific Twitter accounts, smaller and more focused date ranges, particular countries of interest, or specific hashtags, would yield more targeted and meaningful results. Hence, we provide our Python source code to help researchers quickly apply and adapt the model for different use scenarios.

**Sentiment intensity and emotion intensity scores**. The accuracy in determining "valence_intensity", "fear_intensity", "anger_intensity", "happiness_intensity", and "sadness_intensity" are systematically validated in prior research [17] and are subsequently tested for their predictive validities in other NLP tasks [21, 22, 23].

The *descriptive* validity of CrystalFeel is reported in Gupta and Yang's original evaluation experiments using out-of-training-sample test data: the CrystalFeel algorithms' accuracies in terms of Pearson correlation coefficient (*r*) with manually annotated test data are 0.816 on valence intensity, and are 0.708, 0.740, 0.700 and 0.720 on happiness[13] intensity, anger intensity, fear intensity and sadness intensity [17]. The present algorithms we used have reported higher Pearson r values, i.e., 0.765, 0.818, 0.788, 0.765, and 0.856 on predicting fear intensity, anger intensity, happiness intensity, sadness intensity and valence intensity, respectively [16].

The *predictive* validity of the valence, happiness, anger, fear and sadness intensity scores on other tasks have been studied and demonstrated in the context of predicting news social popularity on Facebook and Twitter [21], in predicting the ingredients of happy moments [22], and in detecting propaganda techniques in news articles [23]. Hence, researchers may examine the use of the sentiment and emotion intensity scores directly, instead of their converted, reduced forms of categorical outputs on sentiment categories and emotion categories.

---

[13] In the context of this paper, "joy" and "happiness" are used interchangeably.

**Sentiment and emotion labels.** The "sentiment" and "emotion" attributes are obtained based on a conversion logic presented in the "Methods" section. The conversion principle that allows each tweet to be labelled with one of the five emotion categories (i.e., "fear", "anger", "happiness", "sadness", "no specific emotion") follows a conceptual simplification that a single dominant emotion exists for each tweet.

However, some tweets may express "mixed emotions" [24], such as expressing anger and fear simultaneously. Other conversion logic may be explored in future research. For example, Mohammad et al. [18] suggest using the mid-scale threshold, i.e., 0.5, to differentiate high-intensity vs. non-high-intensity emotions. Researchers shall examine the intended applications and determine the conversion threshold accordingly.

**Topic coverage**
We checked the volume of the tweets related to the top ten identified topic clusters. A vast majority of tweets were related to two or more topics, which form 60% of the total tweets. The tweets that *solely* pertained to "t1" have the highest volume, consisting of 33,680,867 tweets or 25.5% of the total data volume.

Table 7 presents the overall tweet topics statistics. Figure 2 depicts a visualization of the topic clusters in the context of the volume of the tweets. (Note that for the current release, topic data is updated until 1 January 2021).

**Table 7. Overall tweets distribution over topic clusters (28 January 2020 – 1 January 2021)**

| Tweets related to attributes | Count of tweets | Share |
|---|---|---|
| **two or more topics** | 76,863,395 | 58.2% |
| **t1 only** | 33,680,867 | 25.5% |
| **t2 only** | 14,148,720 | 10.7% |
| **t3 only** | 482,1683 | 3.7% |
| **t4 only** | 156,0158 | 1.2% |
| **t5 only** | 218,663 | 0.2% |
| **t6 only** | 136,134 | 0.1% |
| **t7 only** | 122,517 | 0.1% |
| **t8 only** | 89,380 | 0.1% |
| **t9 only** | 18,184 | 0.01% |
| **t10 only** | 77,414 | 0.05% |
| other single topics | 362,495 | 0.2% |
| total | 132,099,160 | 100% |

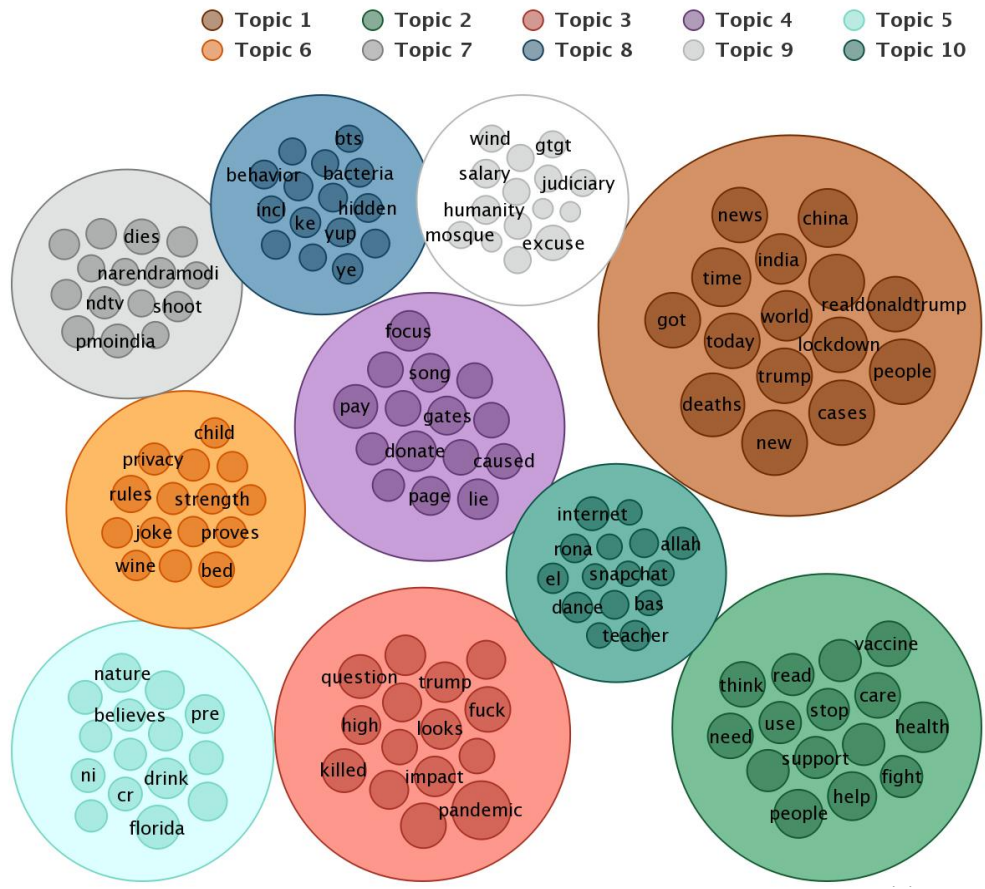

**Figure 2. The ten topic clusters showing different issues surfaced in the total tweets**

**Sentiment coverage**

The quantitative "sentiment_intensity" averaged for the whole COVID-19 dataset is 0.451, with the most negative tweet having its valence intensity score of -0.058, and the most positive tweet having its valence intensity score of 1.005 (Table 8). Vaccine-related tweets data has relatively lower valence intensity.

**Table 8. Sentiment intensity descriptive statistics**

| Attribute | Mean | Standard deviation | Median | Min | Max |
|---|---|---|---|---|---|
| valence_intensity (COVID-19 tweets) | 0.451 | 0.097 | 0.451 | -0.058 | 1.005 |
| valence_intensity (vaccine tweets) | 0.441 | 0.094 | 0.438 | -0.023 | 0.965 |

Qualitatively, the counts and distributions for valence intensity scores converted into sentiment categories counts are presented in Table 9. The results indicate that most of the tweets are "negative" or "very negative", forming 61.2% of the total 252,600,524 tweets.

**Table 9. Sentiments count and share**

| Sentiment category | Count of tweets | Share |
|---:|---:|---:|
| COVID-19 tweets | | |
| very negative | 14,731,331 | 5.5% |
| negative | 140,583,388 | 55.7% |
| neutral | 37,841,749 | 15.0% |
| positive | 57,570,402 | 22.8% |
| very positive | 1,873,654 | 0.7% |
| total | 252,600,524 | 100% |
| Vaccine tweets | | |
| very negative | 842,157 | 6.2% |
| negative | 8,089,530 | 59.9% |
| neutral | 1,878,739 | 13.9% |
| positive | 2,619,128 | 19.4% |
| very positive | 76,183 | 0.6% |
| total | 13,505,737 | 100% |

Plotting the "sentiment" values over *daily* aggregated tweet counts suggested subtler patterns (see Figure 3). For example, the single-day peak during this period was 1,075,087 tweets (629,938 were "negative" tweets and 65,165 were "very negative" tweets), which took place on 13 March 2020, one day immediately following the WHO's announcement on the disease as a "pandemic". Further analysis may look into, for example, the sentiment changes before and after more targeted periods based on critical announcements (e.g., to study a week before and after 13 March 2020). The dataset may also allow further research to explore the correlations and predictive values based on the sentiment and emotion scores when overlaid with economic indicators (e.g., stock market changes).

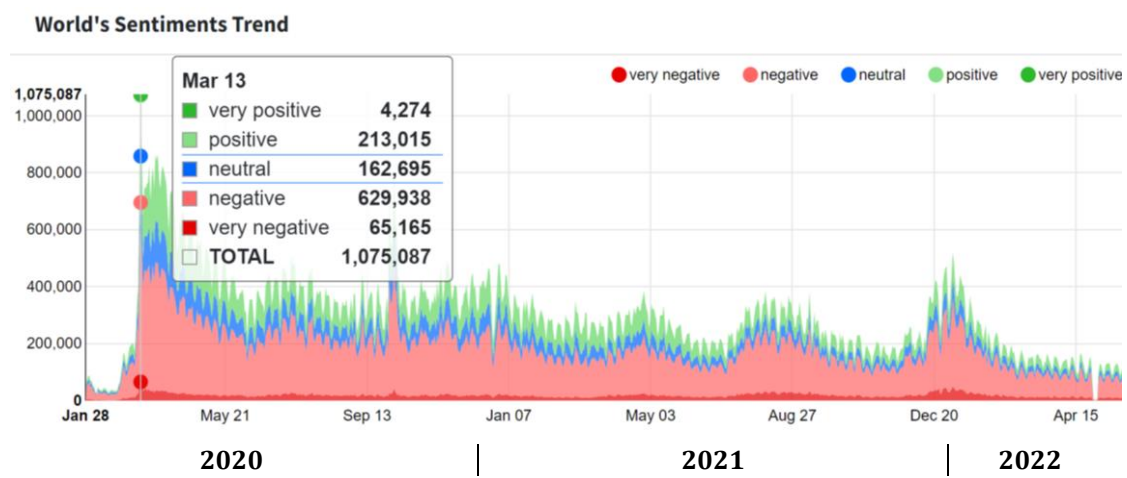

**Figure 3. Five sentiment categories in the stacked chart showing their evolvement over time at the global scale**

**Emotion coverage**
Using the four quantitative emotions intensities attributes, overall statistics show that "anger_intensity" has the highest mean value of 0.450, followed by "fear_intensity" with its intensity value of 0.445. Table 10 reports the descriptive statistics for the four emotion intensity scores.

**Table 10. Emotion intensity descriptive statistics**

| Attribute | Mean | Standard deviation | Median | Min | Max |
|---:|---:|---:|---:|---:|---:|
| COVID-19 tweets | | | | | |
| fear_intensity | 0.445 | 0.097 | 0.448 | -0.053 | 1.198 |
| anger_intensity | 0.450 | 0.091 | 0.445 | -0.007 | 1.102 |
| sadness_intensity | 0.422 | 0.082 | 0.416 | 0.011 | 1.060 |
| happiness_intensity | 0.291 | 0.096 | 0.288 | -0.117 | 0.984 |
| Vaccine tweets | | | | | |
| fear_intensity | 0.471 | 0.089 | 0.471 | -0.022 | 1.072 |
| anger_intensity | 0.466 | 0.091 | 0.463 | 0.048 | 1.006 |
| sadness_intensity | 0.421 | 0.077 | 0.417 | 0.058 | 0.968 |
| happiness_intensity | 0.263 | 0.094 | 0.256 | -0.08 | 0.956 |

Qualitatively, the counts and distribution of the most dominant emotion categories based "emotion" attribute are presented in Table 11. The results suggest that, over the 28 months, tweets that are dominantly expressing "anger" (75,824,459 tweets, 30.0%) and tweets that are dominantly expressing "fear" (64,633,650 tweets, 25.6%) formed the majority of the total tweets. For vaccine-related tweets, the data have a roughly equal share of anger (32.7%) and fear (31.2%) tweets.

**Table 11. Sentiments count and share**

| Sentiment | Count | Share |
|---:|---:|---:|
| COVID-19 tweets | | |
| fear | 64,633,650 | 25.6% |
| anger | 75,824,459 | 30.0% |
| sadness | 14,856,610 | 5.9% |
| happiness | 59,444,056 | 23.5% |
| no specific emotion | 37,841,749 | 15.0% |
| total | 252,600,524 | 100% |
| Vaccine tweets | | |
| fear | 4,217,313 | 31.2% |
| anger | 4,415,281 | 32.7% |
| sadness | 299,093 | 2.2% |
| happiness | 2,695,311 | 20.0 |
| no specific emotion | 1,878,739 | 13.9% |
| total | 13,505,737 | 100% |

We checked the daily counts of the four emotions for the 28 months (see Figure 4). The significance of the change can be illustrated using the contrast of results at the start and the end of our data range.

For example, as of 28 January 2020, a total of 23,405 tweets were posted for the day, and the tweets with "anger" as the dominant emotion formed 15.3% of the total 23,405 tweets, far less than those tweets with "fear" as the most dominant emotion which formed 52.5% of the total 23,405 tweets. In contrast, as of 5 October 2020, a total of 602,206 total tweets were posted for the day, and the tweets with "anger" as the most dominant emotion formed 39.1%, exceeding those tweets with "fear" as the most dominant emotion, which formed 20.1%).

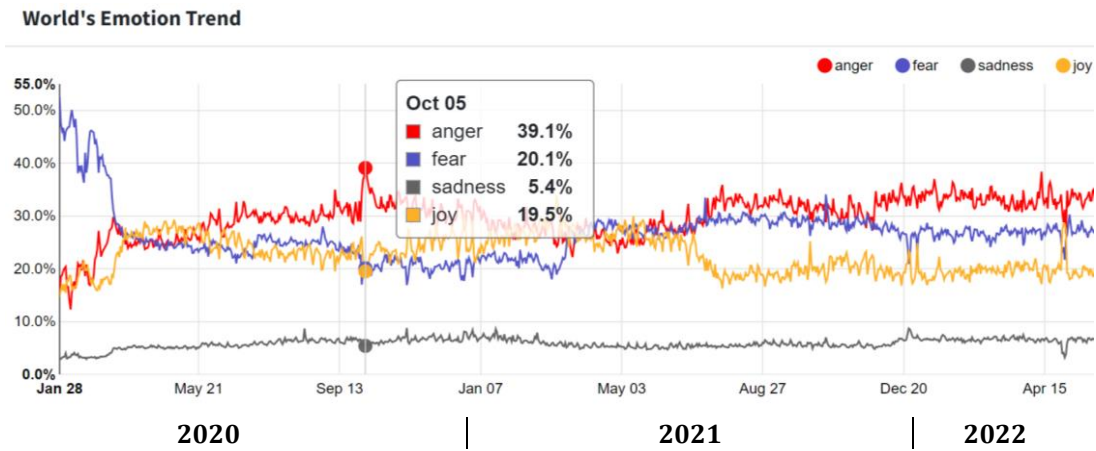

Figure 4. Daily distribution of the four primary emotions over time

The trends surfaced some interesting patterns: While both "fear" and "anger" dominated in the overall counts, the trend plot shows that over time, the relative distribution of "fear" has been decreasing, and the relative distribution of "anger" has been increasing; meanwhile, "happiness and other positive expressions" have been increasing, though in a slower rate (See [7] which provides an interpretation based on analysis of early time coverage of this dataset).

**Geographical representativeness**
We checked the "country/region" attribute converted from the "location" attribute to understand the geographic coverage and representativeness of the dataset. The geographical coverage of the tweets is estimated to contain users coming from more than 170 countries, regions, or territories worldwide. Figure 5 shows a visualization of the dataset's geographical representativeness.

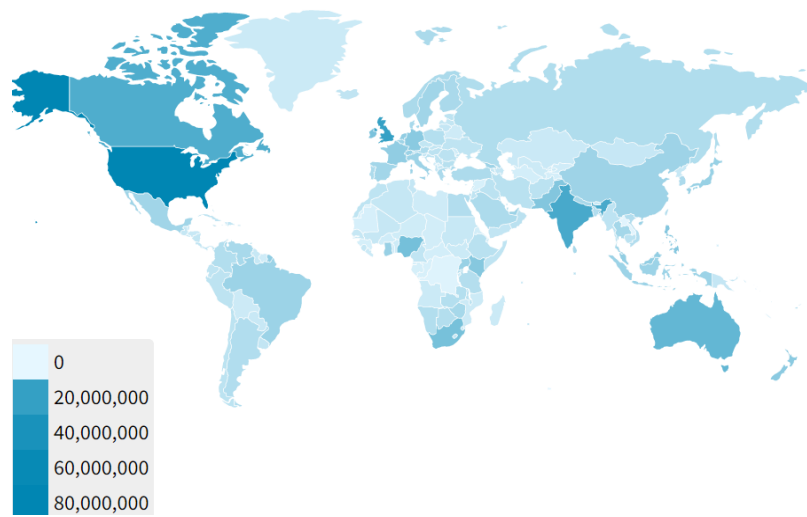

Figure 5. Geographical coverage and volume distribution of the tweets

Table 10 shows the number of tweets that are identifiable for 30 representative countries. The top ten countries that reported the highest number of Coronavirus-related tweets in English are the United States, the United Kingdom, India, Canada, Australia, Nigeria, South Africa, Ireland, Pakistan, and Kenya. According to Statista's estimate [25], the United States, Japan, India, United Kingdom, Brazil, Indonesia, Turkey, Saudi Arabia, Mexico, and France are leading countries based on the number of Twitter users as of January 2021.

**Table 12. Number of tweets breakdown by 30 counties (28 January 2020 – 1 June 2022)**

| Country | WHO Region[14] | Count | Share |
|---|---|---|---|
| Argentina | Americas | 68,645 | 0.03% |
| Australia | Western Pacific | 4,533,585 | 1.8% |
| Brazil | Americas | 305,200 | 0.12% |
| Canada | Americas | 9,049,597 | 3.6% |
| China | Western Pacific | 288,166 | 0.11% |
| Colombia | Americas | 61,246 | 0.02% |
| Denmark | Europe | 135,245 | 0.05% |
| France | Europe | 609,958 | 0.24% |
| Indonesia | South-East Asia | 305,709 | 0.12% |
| India | South-East Asia | 11,148,980 | 4.4% |
| Iran | Eastern Mediterranean | 34,551 | 0.01% |
| Ireland | Europe | 1,771,300 | 0.7% |
| Italy | Europe | 299,455 | 0.12% |
| Japan | Western Pacific | 386,460 | 0.15% |
| Kenya | Africa | 1,052,348 | 0.42% |
| Malaysia | Western Pacific | 571,054 | 0.23% |
| New Zealand | Western Pacific | 823,054 | 0.32% |
| Nigeria | Africa | 2,225,393 | 0.9% |
| Pakistan | South-East Asia | 1,198779 | 0.5% |
| Philippines | Western Pacific | 1,092,815 | 0.4% |
| Russia | Europe | 73,756 | 0.03% |
| Singapore | Western Pacific | 483,706 | 0.19% |
| South Africa | Africa | 2,046,941 | 0.8% |
| South Korea | Western Pacific | 103,459 | 0.04% |
| Spain | Europe | 197,981 | 0.08% |
| Thailand | South-East Asia | 180,352 | 0.07% |
| Turkey | Europe | 96,986 | 0.04% |
| United Kingdom | Europe | 18,736,635 | 7.4% |
| United States | Americas | 68,935,089 | 27.3% |
| Viet Nam | Western Pacific | 66,184 | 0.03% |
| sub-total with country identifiable information | | 134,718,945 | 53.3% |
| sub-total without country identifiable information | | 117,881,579 | 46.7% |
| | total | 252,600,524 | 100% |

Overall, we found that 134,718,945 tweets (53.3% of the 252,600,524 tweets) have country-identifiable "location" information as declared by the users at their Twitter public profile. We believe that this is a reasonable proportion to facilitate assessing the dataset's geographic representativeness.

---

[14] WHO grouped its member states into six regions: Africa, Amaricas, South-East Asia, Europe, Eastern Mediterranean, Western Pacific. Source: https://www.who.int/about/who-we-are/regional-offices

Table 13 shows the number of vaccine-related tweets that are identifiable for ten countries. The top three countries that reported the highest number of vaccine-related tweets in English are the United States, the United Kingdom, and Canada.

**Table 13. Number of vaccine-related tweets breakdown by 10 counties (28 January 2020 – 1 June 2022)**

| Country | Count | Share |
|---|---|---|
| Australia | 284,476 | 2.1% |
| Canada | 667,156 | 4.9% |
| Germany | 39,549 | 0.3% |
| India | 260,857 | 1.9% |
| Ireland | 86,802 | 0.6% |
| Malaysia | 363,71 | 0.3% |
| New Zealand | 73,051 | 0.5% |
| South Africa | 101,523 | 0.8% |
| United Kingdom | 874,520 | 6.5% |
| United States | 3,493,318 | 25.9 |
| sub-total with country identifiable information | 7,043,520 | 52.2% |
| sub-total without country identifiable information | 6,462,217 | 47.8% |
| total | 13,505,737 | 100% |

## Usage Notes

This paper presents a very large COVID-19 Twitter dataset with psychologically meaningful attributes. This dataset may create opportunities to understand both global and local conversations and social sentiments in real-time, at a large scale, potentially leading to rich insights into human behaviours and behavioural changes surrounding the unprecedented pandemic. We discuss its current usage and envisage future use in five broad areas.

**Media and communication research**. First, for media and communication research, the dataset can be helpful for communication scientists and professionals to evaluate and improve government response, policies, and media communications towards the unprecedented pandemic crisis. For example, a recent study compared the communications efforts of health authorities in the United States, the United Kingdom, and Singapore on Facebook during the early period of COVID-19 [26].

Lwin et al. [7] used a part of this dataset and presented the first rapid report that has shed light on the global sentiment change in people's emotional responses to the pandemic. It focused on the emotion category attribute and found that anger has overtaken fear as the dominant emotion in the share of tweets from 28 January 2020 to 9 April 2020. A follow-on study by Lwin et al. [27] examined the evolution of sentiments, using a comparative analysis of change in sentiments across five countries India, Singapore, South Korea, the United Kingdom, and the United States from 28 January 2020 to 28 April 2021. The study focused on the sentiment category attribute as the main index variable and found that while negative sentiments were predominant during the initial period of the pandemic, positive sentiments arose at differing intensities across the five countries, particularly in Asian countries.

As the virus continuously affects different countries in different timeframes and governments implemented different response strategies and policies, one direction of an ongoing related work overlays the location attribute, examining and comparing sentiments, emotions, and topics associated with different countries. The dataset may also help to study how media's topical and emotional framing in their headlines and titles are different from those expressed by the general public.

**Psychology**. Second, the dataset is of inherent interest for psychology research. The granularity of the tweet's metadata may allow researchers to dive deeper into more nuanced trends with profound psychological accounts and insights. One possibility is to look into the sentiments and emotional differences over more fine-grained timelines, examine cultural differences and segregate the users who are influencers vs. the general public. Research may also look into leveraging user characteristics inference techniques (e.g., [28]) and the present dataset to investigate user community-specific tendencies and issues.

Kamila et al. [29] used a part of this dataset and examined the association between one's temporal orientation (focusing on past, present or future) and emotions (fear, anger, joy and sadness) using a part of this dataset. The study revealed that future orientation is associated with *joy* and *anger*, and past orientation is associated with *fear* and *sadness*.

**Population mental health and well-being**. Third, as the pandemic escalates in its severity and geographical span and is likely to last for a prolonged period, public mental health issues (e.g., [30]) are more prevalent. The dataset may be used to examine public mental health and well-being. Prior literature (e.g., [31, 32]) has established the linkage between fear (as an emotion) and anxiety (as a mental disorder) and between sadness (as an emotion category) and depression (as a mental disorder). It can be fruitful to study the value of the emotion intensity scores and their trends in the context of their duration, frequency, and various user communities.

In the area of supporting mental healthcare needs, Nimmi et al. [33] trained a deep learning ensemble model for emotion identification using a part of this dataset and call content of an emergency response support system. They found the new model achieved high accuracy in terms of attaining a macro-average F1 score of 85.2 per cent.

**Economics**. Fourth, it is potentially valuable to overlay publicly available economic indicators (e.g., daily stock market data, monthly unemployment rates reports) and investigate how the topics, sentiments, and emotional trends present predictive value in future research.

**Epidemiology**. Last but not least, data scientists and epidemiology researchers may find the dataset useful. For example, prior research on Zika [34] and other infectious disease outbreaks [35] has studied and found insights into overlaying with air travel networks and virus genome. Hence the dataset may reveal more hidden patterns and relationships between the large-scale social media content and other pandemic-related data streams. Future work may also explore the forecasting or predictive value of the social media data and emotion indicators in the spread of the virus and cases.

## Data Availability

The dataset described in this paper is available for download at Open ICPSR: https://doi.org/10.3886/E120321. The dataset license is also available in the Open ICPSR download folder. Essentially, the dataset license is based on the CC BY-NC 2.0 template and considers the need to be consistent with Twitter's terms of service as the dataset is built upon the content provided by Twitter's standard API.

## Code Availability
The source scripts for the trained LDA-based topic model are available on our GitHub page: https://github.com/ajvish91/covid_twitter_scripts. A visualization dashboard on the COVID-19 tweets with hourly refreshed sentiment and emotion trend results is available at: https://socialanalyticsplus.net/corona2019. CrystalFeel is accessible via: https://socialanalyticsplus.net/crystalfeel. Access to CrystalFeel API is available upon request from the corresponding author.

## Acknowledgements
This work is supported in part by the Agency for Science, Technology and Research (A*STAR), Singapore under its A*ccelerate Gap Fund (ETPL/18-GAP050-R20A) and SERC Council Strategic Fund (C210415006), and in part by the Singapore Ministry of Health's National Medical Research Council under its COVID-19 Research Fund (COVID19RF-005 & COVID19RF-0009).

We thank Andrew Ortony, Anita Sheldenkar, Brandon Loh, Cui Mengyang, Gangeshwar Krishnamurthy, Lim Keng Hui, Jiahui Lu, Joseph JP Simons, May Oo Lwin, Nur Atiqah Othman, Paul E. Cain, Sebastian Maurer-Stroh, Therese Quieta, Thomas Vuong, Tuan Le Mau, Wan Kum Seong, Wong Chi Kit, and Zhang Mila, for helpful discussions. We are grateful for the help from Nur Atiqah Othman for her proofreading, which helped to enhance the clarity of the paper. All errors that remain are our sole responsibility.

## Author Contributions
RG acquired the data and extracted sentiment and emotion features. AV extracted topic cluster features. YY initiated, conceptualized, and led the manuscript. All authors performed data analysis, contributed to the manuscript writing, reviewed the content, and agreed with the submission.

## Competing Interests
The authors declare the following competing interests: RG and YY are co-inventors of the CrystalFeel tool used to extract the sentiment and emotion-related attributes. No other conditions or circumstances present a potential conflict of or competing interest for the other authors.

22. Gupta, R.K., Bhattacharya, P. & Yang, Y. (2019). What constitutes happiness? Predicting and characterizing the ingredients of happiness using emotion intensity analysis, In *Proceedings of the AAAI-19 Workshop on Affective Content Analysis (AAAI-AFFCON)*. http://ceur-ws.org/Vol-2328/4_3_paper_22.pdf
23. Krishnamurthy, G., Gupta, R.K. & Yang, Y. (2020). SocCogCom at SemEval-2020 Task 11: Characterizing and Detecting Propaganda Using Sentence-Level Emotional Salience Features, In *Proceedings of the 14th International Workshop on Semantic Evaluation (COLING-SemEval)*. https://doi.org/10.18653/v1/2020.semeval-1.235
24. Berrios, R., Totterdell, P., & Kellett, S. (2015). Eliciting mixed emotions: a meta-analysis comparing models, types, and measures. *Frontiers in Psychology*, 6, 428. https://doi.org/10.3389/fpsyg.2015.00428
25. Statista, *Leading countries based on number of Twitter users as of January 2021(in millions)*. https://www.statista.com/statistics/242606/number-of-active-twitter-users-in-selected-countries (Accessed on 11 February 2021)
26. Sesagiri Raamkumar A, Tan SG & Wee HL. (2020). Measuring the Outreach Efforts of Public Health Authorities and the Public Response on Facebook During the COVID-19 Pandemic in Early 2020: Cross-Country Comparison. . *Journal of Medical Internet Research. 22.* https://doi.org/10.2196/19334
27. Lwin, M.O., Sheldenkar, A., Lu, J., Schulz, P.J., Shin, W., Panchapakesan, C., Gupta, R., and Yang, Y. (2022). The evolution of public sentiments during the COVID-19 pandemic: Case comparisons of India, Singapore, South Korea, the United Kingdom and the United States, *JMIR Infodemiology*. 2(1):e31473. https://doi.org/10.2196/31473
28. Volkova, S., Bachrach, Y., Armstrong, M. & Sharma, V., (2015). Inferring latent user properties from texts published in social media. In *Proceedings of Twenty-Ninth AAAI Conference on Artificial Intelligence*. https://dl.acm.org/doi/10.5555/2888116.2888374
29. Kamila, S., Hasanuzzaman, M., Ekbal, A. et al. (2022). Investigating the impact of emotion on temporal orientation in a deep multitask setting. *Scientific Reports.* 12. https://doi.org/10.1038/s41598-021-04331-3
30. Luo, M., Guo, L., Yu, M. & Wang, H. (2020). The Psychological and Mental Impact of Coronavirus Disease 2019 (COVID-19) on Medical Staff and General Public – A Systematic Review and Meta-analysis, *Psychiatry Research.* 291. https://doi.org/10.1016/j.psychres.2020.113190
31. Öhman, A., (2008). Fear and anxiety. *Handbook of Emotions*, 709-729.
32. Karp, D.A., (2017). *Speaking of sadness: Depression, disconnection, and the meanings of illness*. (Oxford University Press, 2017).
33. Nimmi, K., Janet, B., Kalai Selvan, A., Sivakumaran, N. (2022). Pre-trained ensemble model for identification of emotion during COVID-19 based on emergency response support system dataset, *Applied Soft Computing* 122. https://doi.org/10.1016/j.asoc.2022.108842
34. Grubaugh ND, Saraf S, Gangavarapu K, et al. (2019). Travel Surveillance and Genomics Uncover a Hidden Zika Outbreak during the Waning Epidemic, *Cell* 178 1057-1071. https://doi.org/10.1016/j.cell.2019.07.018
35. Tatem AJ, Rogers DJ & Hay SI. (2006). Global transport networks and infectious disease spread. *Advances in Parasitology*. 62. 293–343. https://doi.org/10.1016/s0065-308x(05)62009-x
23